\documentclass[,referee,twocolumn,final,authoryear]{elsarticle}

\usepackage{ycviu}
\usepackage{framed,multirow}

\usepackage{amssymb}
\usepackage{latexsym}

\usepackage{lineno,hyperref}
\modulolinenumbers[5]
\usepackage{hyperref}
\usepackage{multicol}
\usepackage{graphicx}
\usepackage{xcolor,cancel,mathrsfs,amscd}
\usepackage{subfig}
\usepackage{amsmath}
\usepackage{amssymb}
\usepackage{amsfonts}
\usepackage{multirow}
\usepackage{rotating}
\usepackage{subfig}
\usepackage{color}
\usepackage{float}
\usepackage{url}
\usepackage{breakcites}
\usepackage{array}

\newcolumntype{L}{>{\centering\arraybackslash}m{3cm}}

\usepackage[nameinlink,capitalize]{cleveref}

\usepackage[T1]{fontenc}
\usepackage{longtable}
\usepackage{supertabular}
\usepackage{hyperref}
\usepackage{xcolor}
\usepackage[graphicx]{realboxes}
\usepackage{pdflscape}
\usepackage{amsmath}
\usepackage{ulem}
\usepackage[edges]{forest}

\usepackage[cmintegrals]{newtxmath}
\definecolor{LightCyan}{rgb}{0.88,1,1}
\usepackage{multirow}
\usepackage{geometry}
\usepackage[edges]{forest}

\definecolor{myblue}{rgb}{.0,.0,.8}

\definecolor{myg}{rgb}{.0,.3,0}

\definecolor{normal}{rgb}{.0,.0,.0}
\definecolor{myred}{rgb}{.8,.0,.0}
\definecolor{myblue}{rgb}{.0,.0,.8}

\usepackage{url}
\usepackage{xcolor}
\definecolor{newcolor}{rgb}{.8,.349,.1}

\journal{Computer Vision and Image Understanding}

\begin{document}

\begin{frontmatter}

\title{High-level Prior-based Loss Functions for Medical Image Segmentation: A Survey}

\author[1,3]{Rosana \snm{El Jurdi}} 
\ead{rosana.el-jurdi@univ-rouen.fr}
\author[1]{Caroline  \snm{Petitjean}}
\ead{caroline.petitjean@univ-rouen.fr}
\author[1]{Paul \snm{Honeine}}
\ead{paul.honeine@univ-rouen.fr}
\author[2]{Veronika \snm{Cheplygina}}
\ead{v.cheplygina@tue.nl}
\author[3,4]{Fahed \snm{Abdallah}}
\ead{fahed.abdallah76@gmail.com}
\address[1]{Normandie Univ, INSA Rouen, UNIROUEN, UNIHAVRE, LITIS, Rouen,76800, France}
\address[2]{Medical Image Analysis group, Eindhoven University of Technology, Eindhoven, The Netherlands}
\address[3]{Universit\'e Libanaise, Hadath, Beyrouth, Liban}
\address[4]{ICD, M2S, Universit\'e de technologie de Troyes, Troyes, France}

\begin{abstract}
Today, deep convolutional neural networks (CNNs) have demonstrated state of the art performance for supervised medical image segmentation, across various imaging modalities and tasks. Despite early success, segmentation networks may still generate anatomically aberrant segmentations, with holes or inaccuracies near the object boundaries. To mitigate this effect, recent research works have focused on  incorporating spatial information or prior knowledge to enforce anatomically plausible segmentation. If the integration of prior knowledge in image segmentation is not a new topic in classical optimization approaches, it is today an increasing trend in CNN based image segmentation, as shown by the growing literature on the topic. In this survey, we focus on high level prior, embedded at the loss function level. We categorize the articles according to the nature of the prior: the object shape, size, topology, and the inter-regions constraints. We highlight strengths and limitations of current approaches, discuss the challenge related to the design and the integration of prior-based losses, and the optimization strategies, and draw future research directions.

\end{abstract}


\end{frontmatter}
\begin{figure*}[t!]
  \subfloat[][From left to right: Neuronal image, ground truth, baseline model segmentation, segmentation with prior]
  {
	\begin{minipage}[c]{
	   0.475\textwidth}
	   \centering
	   \includegraphics[width=8cm,height=2cm]{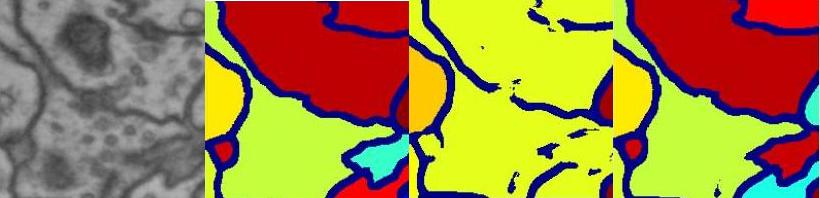} \label{fig:NeuronPLausibility} 
	    \\
	\end{minipage}
	}
 \hfill 	
  \subfloat[][From left to right: ground truth segmentation of a brain lesion in MRI, baseline segmentation, prior segmentation]
  {
	\begin{minipage}[c]{
	   0.475\textwidth}
	   \centering
	   \includegraphics[width=8cm,height=2cm]{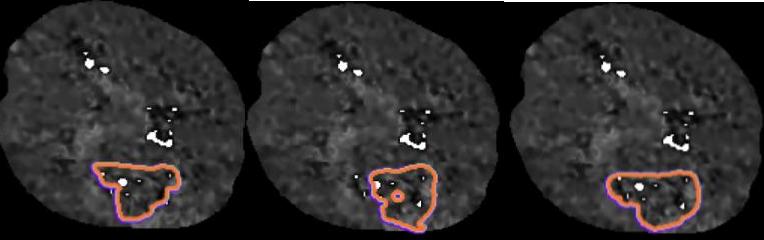} \label{fig:LesionPLausibility}  
	\end{minipage}
	}
\caption{Example illustrations on the relevance of incorporating topological priors in neuronal  (a) and  brain lesion segmentation (b). In (a), segments obtained by baseline method with no topological prior shows broken membranes and incorrect grouping of neuronal areas contrary to prior model that preserves these topological particularities. In (b), baseline segmentation has prohibited holes which are evaded the segmentation with prior case. Figures from \citep{Hu2019} (a) and \citep{Kervadec2019_BoundaryLoss} (b). \textit{[permission pending]}} \label{Anatomic_plausibility}
\end{figure*}
\section{Introduction}
\label{sec:introduction}

Medical image segmentation can be viewed as the process of making per-pixel predictions in an image, in order to identify organs or lesions from the background. Imaging modalities include magnetic resonance imaging (MRI), computed tomography (CT), nuclear medicine functional imaging, ultrasound imaging, fundus photography, to name a few.  The results of segmentation can be used to compute biomarkers or quantitative measurements, to compute three-dimensional anatomical models for image-guided surgery, and to design the radiation beam in radiotherapy planning, in order to spare healthy organs while intensifying the beam on the tumor. As such, the segmentation task is a key step in assisting early disease detection, diagnosis, monitoring treatment and follow up. 

Due to the recent advances in deep learning, computer vision tasks, including automated image segmentation, have experienced a major breakthrough. The main reason for the success of convolutional neural networks (CNNs) lies in their ability to learn hierarchical feature representations automatically through training directly from images, eliminating the need for handcrafted features. State-of-the-art architectures for image segmentation typically have an encoder-decoder structure \citep{Long2015,Unet}, that allows for an end-to-end processing. 
One of the  key component of CNN training is the loss function, as it drives the back-propagation of the error between the predicted value and the reference label. In segmentation, i.e. two mainly used losses are the cross-entropy and the Dice loss function, which is a metric-inspired loss, based on a soft approximation of the Dice score  \citep{V-NET}. 
Still, despite early success, segmentation networks may generate anatomically aberrant segmentations, with holes, voids or high inaccuracies close to organ boundaries (\figurename s~\ref{fig:NeuronPLausibility} and ~\ref{fig:LesionPLausibility}). For example, the winner of the ACDC challenge, which focused on the automatic delineation of the cardiac structures, was an ensemble of U-Net inspired architectures proposed by \cite{isensee_automatic_2018}. Even though this approach obtained the top accuracy, it was not able to prevent aberrant and anatomically impossible segmentation, in 41 patients out of 50 of the testing database \citep{ACDC}.

Embedding prior knowledge on the object, such as shape, appearance or location, into the deep learning networks could allow the increase of network robustness and accuracy, while generating anatomically plausible segmentation, as an increasing number of research papers on the topic demonstrate. Prior knowledge is indeed available in the medical domain, due to conventions in patient position and similarity shared in body structure. It has been commonly exploited in variational approaches before deep learning \citep{Nosrati2016} and is now also a growing direction with regard to CNNs. Prior-based loss functions are also helpful in weakly supervised settings, where only partial labels (e.g. scribbles or bounding boxes) are available \citep{ELJURDI2019_GRETSI,ELJURDI2020_BBUNET,kervadec2020}.

In order to incorporate prior knowledge in a segmentation process, two major questions arise. First, one needs to define the type of prior information and the modeling of the prior. The term "prior knowledge" is quite vague and covers a breadth of notions, as can be collected from the literature. It may refer to spatial or context information, under the form of distance maps or the image gradient, to more complex knowledge, regarding the anatomy of the object of interest (such as shape and size for example) and the connectivity between regions.

Second, one needs to specify how to embed the prior into the segmentation network. While the network architecture can be modified to integrate priors, another way to incorporate prior into the segmentation frameworks is at the level of the loss function. The loss function indeed offers a versatile way to include anatomical constraints at different scales, while maintaining interactions between regions as well as the computational efficiency of the backbone network.  However, designing novel losses for CNN-based segmentation poses several specific challenges: e.g. how to translate the desired anatomical properties on the network output, which is a real-valued probability map, or how to insure differentiability or convexity of the designed prior-based loss.

The goal of this paper is to establish an overview of recent contributions that focus on incorporating prior knowledge at the loss function level, in deep networks for medical image segmentation. We want to add to the understanding of the mechanisms and intuition behind the design and implementation of prior based losses.
In our survey, we do not intend to cover all types of prior, but rather focus on "high-level" priors. We define high-level prior as features extracted from the object that can help in characterizing  and interpreting it, by opposition to low level operator, such as gradient or distance map.
We have categorized the articles according to the nature of the prior, that may be the object shape, size, topology, and the inter-regions constraints. We seek to provide the reader with (i) what types of priors exist in the literature, how they are modeled and embedded into the loss function, (ii) the major challenges linked to the design of such prior-based losses, and (iii) their common training and optimization strategies.
To do so, Google Scholar was queried for peer-reviewed publications that included tags such as constraint losses, medical image segmentation, prior-based losses, constraint neural networks and anatomical constraints. The papers were then filtered in terms of employing a deep CNN-backbone for segmentation and the novelty present in the design of a new prior loss function.

The rest of the paper is organized as follows. We first review existing surveys in CNN-based medical image segmentation, and provide a short review of prior-based approaches; in the pre-deep learning era in Section \ref{Existing surveys}. Section \ref{fundementals} introduces the fundamentals of deep learning for medical image segmentation. Before diving into the heart of our survey, we briefly review in Section \ref{lowlevel} approaches that make use of low-level prior information, to enforce segmentation consistency, so as to position them with respect to the scope of this survey. Section \ref{categorization} is the proposed categorization of prior-based loss functions along with a review of the corresponding works. Section \ref{challenges} exhibit the common challenges and weaknesses faced while designing such losses and discuss some future trends and perspectives. Finally, we conclude the survey in Section \ref{conclusion}.

\section{Related work} \label{Existing surveys}


\subsection{Surveys in medical image segmentation}
Since the rise of convolutional neural networks in computer vision, various medical image segmentation surveys have been published \citep{Hesamian2019,Jiang2017,Haq2020,Razzak2018}. \cite{Razzak2018} presented a survey of medical image segmentation with deep learning and categorized methods in terms of convolutional neural network structure and training techniques (weakly-, semi- or fully-supervised). \cite{taghanaki_deep_2019} reported contributions in image segmentation for natural and medical images and categorized existing works according to six main categories: model architecture, image modality, loss functions, model types, supervision strategies and multi-task learning. \cite{Haq2020} presented an overview of the different deep learning methods used within the biomedical image segmentation domain and categorized them according to the image modality, the segmentation architectures, and their evaluation metrics. Domain specific reviews also exist, focusing on a particular pathology or organ, such as \cite{Havaei2016}, that presents contributions for brain pathology segmentation, in which papers are categorized according to model characteristics and architectures, training methodology, as well as selected topics such as data augmentation, transfer learning, and incomplete predictions. 


As far as prior based loss functions are concerned, \cite{taghanaki_deep_2019} briefly reviews a few works that integrate boundary and topological prior-based losses, in addition to presenting common loss functions, as well as their variants and combinations. Some prior-embedded losses are described in \cite{Cardiac2020}; however, this survey focuses only on cardiac image segmentation. Thus, to the best of our knowledge, no survey deals specifically with prior-based loss functions in image segmentation.


\subsection{Prior-based segmentation approaches in the pre-deep learning era}\label{preDL}

Among the segmentation methods that existed before the deep learning tsunami, optimization-based approaches have been hugely popular, due to their versatility and efficiency. They consist in obtaining the segmentation by optimizing an appropriate energy functional. In the case where the image domain is considered to be continuous, optimization-based approaches have embraced active contours, level sets, and deformable models in general \citep{xu2000image}. On the other hand, spatially discrete approaches consider the image  as a graph, and include the well-known graph cut and normalized cuts approaches \citep{shi2000normalized,boykov2006graph}, to name a few.

In order to counteract the effect of the noise, occlusion and low contrast in medical images, and to increase anatomical plausibility, researchers used prior information to guide optimization-based segmentation algorithms, well before the advent of deep learning for computer vision. Indeed, optimization-based approaches allow to encode easily some constraints on the segmentation results. Independently from the spatial domain (whether it is spatial or continuous), the energy functional to be minimized comprises at least two terms: a data-fidelity term, related to the image to be segmented, and a regularization term that controls the appearance of the contour, e.g. smoothness. One convenient way to add prior information is to embed it into an additional term in the objective functional or via a probabilistic formulation. The additional term contains a dissimilarity measure between the model feature and the segmented region feature.
Depending on the targeted property, the additional term is designed similarly to the data term or to the smoothness term.

Prior encompass a breadth of various forms, as distinguished by \cite{Nosrati2016} in their thoroughgoing review on the topic: they can be based on elementary image properties, such as intensity, color, and texture information, or more elaborate features on the object shape, such as topological and geometrical constraints \citep{vicente08graph}, moment priors \citep{ayed2008area,foulonneau04moments}, distance and adjacency constraints \citep{liu2008graph}, as well as motion and model/atlas-based priors. 
In our review, we will encounter types of prior which are similar to these. Thus, we believe that these past research works could be fruitfully explored to design new losses. For example, \cite{Mirikharaji2018} took inspiration from \cite{olga} to design a star-shape prior.  In the same context, the shape template in
\citep{slabaugh05graph,rousson02shape} or the popular statistical models in \citep{cootes95active,heimann09statistical}, are build based on ground truth segmentation maps and their corresponding  images. These types of priors are not present in our review, since  the CNN is trained with this type of data, and already learns  some appearance and semantic information. For further information on the topic of prior information in optimization bases segmentation approaches, in addition to \cite{Nosrati2016}, we refer the reader to \cite{cremers07review} on shape prior segmentation in variational continuous frameworks, and \cite{grady2012targeted} for spatially discrete frameworks.

In order to be easily optimized, newly-designed energy terms have to be convex (or submodular in spatially discrete frameworks). Interestingly, \cite{Nosrati2016} highlights the  trade-off between the richness of the energy functional, and its optimizability in optimization-based approaches: an accurate modelling of the underlying property will require highly complex or costly optimization. In deep segmentation networks, designing new loss functions includes specific challenges (developed in Section \ref{challenges}) that are not the same as the ones in optimization-based approaches, but our survey could also relate to the fidelity-optimizability trade-off in the reviewed papers.

\section{Fundamentals of medical image segmentation with deep networks} \label{fundementals}

The implementation of medical image segmentation network requires to choose an appropriate architecture, and to train it so as to fix the network weights. Training is done by optimizing a loss function that models the problem at hand, here, pixel labeling. This implementation contains many degrees of freedom, among which include the architecture and the loss function. We will briefly review below state-of-the-art choices in medical image segmentation, for both of them.

\subsection{Segmentation networks}


One of the first CNN architectures to allow automatic end-to-end semantic segmentation is the fully convolutional network (FCN) \citep{Long2015}. The FCN has a structure that is derived from a deep classification model, such as VGG16, AlexNet or GoogLeNet, by removing the corresponding classification layers, i.e. replacing their fully connected layers with convolutional ones, and plugging in an upsampling path that is dedicated to transforming coarse outputs into dense predictions. With its ability to extract multiscale features, FCN has set a milestone in segmentation approaches and paved the way for encoder-decoder segmentation networks.
However, FCN often fails to consider global spatial information, and often result in fuzzy coarse-grained predictions \citep{Ravishankar2017}. Thus, it has given rise to many improved variants, among which U-Net.


The U-Net \citep{Unet} has an encoder/decoder structure, which has the particularity of being symmetric and having skip connections. The encoder part is a contracting path composed of stacked convolutional and max pooling layers, that are dedicated to capturing contextual information in order to detect objects present in the image. The decoder part is an expanding path composed of deconvolutions, also called transposed convolution, or bilinear upsampling layers, that help precise localization of patterns including contours and boundaries. As an image moves further into the contracting layers, it decreases in size but increases in depth of its learnt contextual features. In contrast, the decoder layers increase its input size but decreases its depth until it reaches its initial resolution, thus producing a segmentation map of similar dimensions to that of the input. To make use of both contextual and positional features, skip connections between the downsampling (encoder) and upsampling (decoder) paths are added. Skip connections concatenate symmetrically features from opposing convolution and deconvolution layers. Several variants of U-Net consist in changing the backbone model used for encoding, e.g. VGG and DenseNet, replacing deconvolution layers with super-resolution ones for more concise localization ability, or enhancing the architecture with modifications such as attention mechanisms, dense or residual connections. Since its publication in 2015, the U-Net quickly became a state-of-the-art architecture for medical image segmentation \citep{isensee2019nnu}.

In particular, extensions to 3D images have been proposed in the 3D-UNet model \citep{3D-UNET} and the V-Net model \citep{V-NET}. It is worth noting that several attempts have been made to revisit the U-Net by integrating some prior knowledge into the architecture, such as in \citep{ACNN,ELJURDI2020_BBUNET}; however, such architectural modifications are difficult to engineer with limited or specific use. They are not as flexible as injecting the prior at the loss function level, whose versatility comes from the plug-into-any-CNN-backbone property.

\subsection{Loss functions for image segmentation}\label{lossfunctions}

Given a training dataset of $N$ images $\{\mathbf{x}_i\}_{i=1}^N$ and their corresponding ground truth masks $\{\mathbf{y}_i\}_{i=1}^N$, the goal is to train the segmentation network so that it can learn to approximate the ``true'' function $f$ parameterized by $\theta$, that maps the input image $\{\mathbf{x}\}$ to the predicted label map, i.e. such that $f(\mathbf{x},\theta)$ represents a map with the label probability at each pixel. 
In the following, we will use $\mathbf{\widehat{y}}_i = f(\mathbf{x}_i,\theta)$ and rely on Table \ref{tab:notations} for mathematical notations.

\begin{table*}[t]
\centering
\caption{Mathematical notations}
\label{tab:notations}
\begin{tabular}{cl}
 $ \Omega $ & spatial image domain, $\Omega \subset \mathbb{R}^2$ or $\mathbb{R}^3$ \\ 
 $\mathcal{N}_p$& set of neighboring pixels of pixel $p$ \\
$K$ & number of classes\\ 
$\mathbf{y}$ & true label map of dimension $ |\Omega| $   \\ 
$\mathbf{\widehat{y}} $ & predicted label (probability) map of dimension $ |\Omega|$ \\
$\widehat{y}_p$ & probability of pixel $p$ of having label $y_p$
\\ 
$y_p$& true label of pixel $p$\\ 
$y_p^r$&  binary value indicating whether pixel $p$ belongs to class $r$ or not\\ 
\end{tabular}
\end{table*}


Training the network boils down to finding the network parameters $\theta$ that minimize a loss function $\mathcal{L}(\theta)$. For sake of simplicity, we will drop the dependency in $\theta$ and denote by $\mathcal{L}$ the loss function. The loss function $\mathcal{L}$ reflects the problem at hand, i.e. is a data-fitting loss, that we note as $ \mathcal{L}_{fit}$. It is of the form:
\begin{equation}
    \mathcal{L}_{fit} =\sum_{p \in \Omega} L(\widehat{y}_p, y_p) 
    \label{general}
\end{equation}
where $L$ is a function that penalizes the discrepancy between the predicted pixel label $(\widehat{y}_p)$ and the ground truth label ($y_p$) for each pixel $p \in \Omega$. 
The shape of $L$ defines how the error is computed. It is mainly derived from common norms, such as the $||.||_2$ or the log-norm (cross entropy shape) for example.
When based on a norm, the loss is continuous and differentiable, which allows it to be efficiently optimized during back-propagation. Properties of the loss shape are important to translate task specifications. Symmetry, for instance, ensures equal penalization between errors caused by over-segmentations and ones that are caused by under-segmentations \citep{ICML-2019-SymmetricalLosses}.

The standard segmentation losses are the cross-entropy \citep{Unet} and the soft approximation of the Dice score \citep{V-NET}.
The cross-entropy $ L_{CE}$ is a widely used standard loss function that is formulated via the Kullback–Leibler divergence and computes the dissimilarity between the predicted probabilistic distribution and its corresponding target binary distribution. 
Its mathematical expression given in the case of $K$ classes, is:
\begin{equation}
    \mathcal{L}_{CE} = -\frac{1}{|\Omega|}\sum\limits_{p \in \Omega}\sum_{r = 1}^{K} y^r_p \log(\widehat{y}^r_p).
    \label{cross-entropy}
\end{equation}
Since each pixel is handled independently from its neighbors, problems may arise due to class imbalance, as training can be dominated by the most prevalent class. For this reason, multiple works proposed variants of cross-entropy, with weights according to class or pixel imbalance \citep{Jang2018,Baum2017}. One important cross-entropy variant is the weighted cross-entropy \citep{Unet}, which tackles the cross-entropy sensitivity towards class distributions. Denoted by $L_{WCE}$, it is defined as:
\begin{equation}
   \mathcal{L}_{WCE} = -\frac{1}{|\Omega|} \sum_{r = 1}^{K} \sum\limits_{p \in \Omega} w_k y_p^r \log(\widehat{y}^r_p),
    \label{weightcrossentropy}
\end{equation}
where 
the weighting factor $w_i$ assigns more weight to recessive classes, thus enforcing a higher penalty on their corresponding errors.
Another variant is the focal loss  \citep{Focalloss_2017}, which extends upon cross-entropy in order to deal with the extreme foreground-background class imbalance in images.

Introduced by \cite{V-NET}, the Dice loss $ L_{Dice}$ is a soft approximation of the well-known Dice metric, which penalizes the overlap mismatch between the predicted segmentation map and the corresponding target map. 
It can be computed in the general case with $K$ classes \citep{Sudre2017}:
%
%
%
%
\begin{equation}
  \mathcal{L}_{Dice} = 1 - 2 \frac{\sum\limits_{r = 1}^{K} \sum\limits_{p \in \Omega}y^r_p \widehat{y}^r_p}{ \sum\limits_{r = 1}^{K} \sum\limits_{p \in \Omega}(y^r_p)^2 + \sum\limits_{r = 1}^{K} \sum\limits_{p \in \Omega}(\widehat{y}^r_p)^2}.
\label{dice-loss}
\end{equation}
The Dice loss is sensitive to segmentation errors when the segmented object is small. For this reason, some works have aimed at weighting the Dice loss \citep{Yang2018, Sudre2017} in order to take into account the class imbalance, or extending upon it by accounting for background pixels, such as the Kappa coefficient inspired loss  \citep{zhang20}. 


\subsection{Limitations of U-Net predictions}
The cross-entropy and Dice losses, as well as their variants, and their combinations, are widely used in segmentation. However, these losses ignore high-level features or structures concerning the object of interest, such as their shape or topology. They also penalize all mistakes equally, regardless of their nature. In the same spirit, the U-Net vanilla architecture does not leverage specific, anatomical or contextual constraints, nor does it exploit spatial relationships between organs.  This is why many research works have explored the possibility of introducing prior information. 

Incorporating prior information into segmentation encompasses a wealth of notions and covers various implementations; it can consist in introducing architecture modifications, adding constraints in the optimization problem, adding penalty terms in the loss function, or combining all these modifications. In the next section, Section \ref{lowlevel}, we review some approaches that take benefit from low-level prior, i.e. architectural constraints or extracting features from the label maps, without making use of high-level knowledge in the loss function. Then, in Section \ref{categorization}, we review high-level based priors, that are integrated into the loss term, which is the core of our review. 
\section{Incorporating low-level prior in image segmentation} \label{lowlevel}


\subsection{Feature extraction from label maps}
\label{GTtransformationlosses}
One way to improve segmentation consistency is through using reformulated ground-truth representations. Such transformations are able to reveal geometric and shape attributes through feature extraction \citep{Kervadec2019_BoundaryLoss, caliva2019_MIDL,Arif2018}.
\cite{yang2019_MIDL} exploit Laplacian filters in order to develop a boundary enhanced loss term that invokes the network to generate strong responses around boundary areas while producing a zero response in pixels that are at the periphery.  Distance maps are also helpful, as a penalty-term \citep{caliva2019_MIDL,Kervadec2019_BoundaryLoss} or added into the softmax at the end of the network \citep{petit_biasing_nodate}.
In the same vein, \cite{Arif2018} introduce a shape aware loss function that constraints predictions to conform to permissible manifold in vertebrae segmentation. Different from the previous approaches, \cite{Mosinska2018_CVPR} propose to leverage the topological information or shape descriptors present within the internal layers of VGG16 network, in order to close small gaps in thin structures (neuronal membranes) and  alleviate topology mistakes. \cite{kim} introduce a loss term inspired by the Mumford-Shah functional in order to force each region to have similar pixel values, and to regularize the contour length.  \cite{liu2020deep}  interpret the softmax activation function as a dual variable in a variational problem, and are thus able to impose many spatial priors in the dual space, such as spatial regularization, volume and star-shape priors.

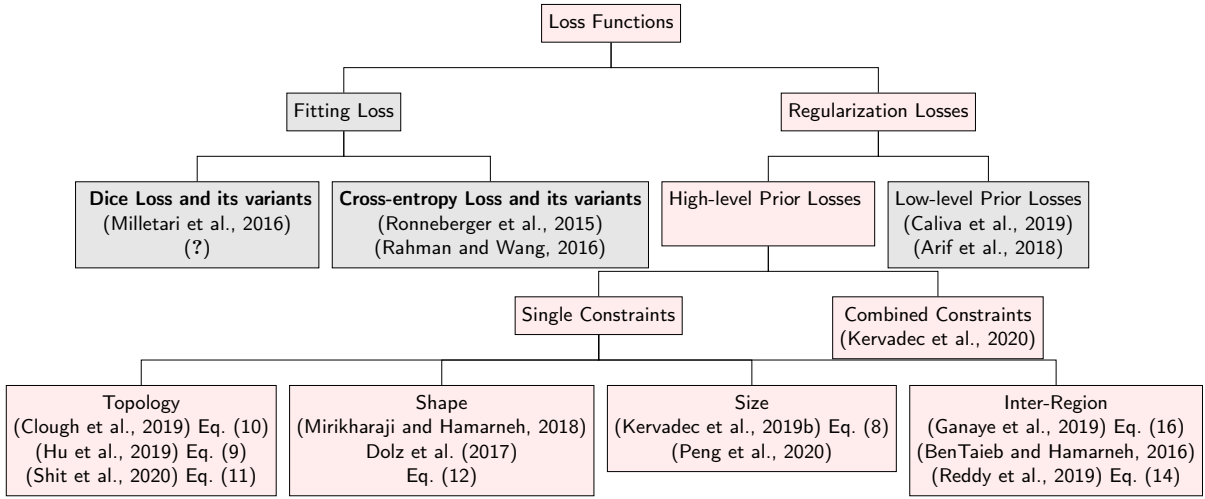
\begin{figure*}[t]
\begin{center} 
\small
\resizebox{0.99\textwidth}{!}{
    \begin{forest}
    for tree={          
         font = \sffamily\linespread{1}\selectfont,
        align = center,
         fill = gray!20, draw,
         grow = south,
    forked edge,        
        s sep = 2mm,    
        l sep = 8mm,    
     fork sep = 4mm,    
               }
[Loss Functions, fill=pink!30
    [Fitting Loss
        [\textbf{ Dice Loss and its variants }\\    \citep{V-NET}\\  \citep{Focal_Loss}, fill=gray!20]
         [\textbf{Cross-entropy Loss and its variants} \\    \citep{Unet} \\   \citep{Rahman2016}]
    ]
    [Regularization Losses, fill=pink!30
        [High-level Prior Losses \bigskip, fill=pink!30,grow=south,
            [Single Constraints, fill=pink!30
                [Topology \\
                \citep{Clough2019} \cref{topo_loss} \\ 
                \citep{hu_NIPS2019}  \cref{Topo_PersistentHomology}\\ 
                \citep{clDice2020} \cref{skeleton-loss}, fill=pink!30]
                [Shape \\ \citep{Mirikharaji2018} \\ \cite{Dolz2017} \\ 
                \cref{Star_shape_prior}, fill=pink!30]
                [Size  \\
                \citep{KERVADEC2019_SizeLoss} \cref{eq:size-loss}
                \\
                \citep{Peng2020_Discrete} , fill=pink!30]
                [Inter-Region \\ 
                \citep{Ganaye2019} \cref{ganaye} \\
                \citep{BenTaieb2016} \\
                \citep{Reddy2019} \cref{inter-1}, fill=pink!30]
            ]
            [Combined Constraints\\
            \citep{kervadec2020},fill=pink!30]
        ]
        [Low-level Prior Losses \\ \citep{caliva2019_MIDL} \\ \citep{Arif2018} , fill=gray!20]
    ]
]
  \end{forest}}
\end{center}
\caption{Loss functions categorization tree. Loss functions can be data-fitting loss or regularization loss. Regularization or prior-based losses can be distinguished according to the nature of prior that they incorporate: low-level prior (Section \ref{GTtransformationlosses}) or high-level prior (Section \ref{categorization}). For each paper, the equation refers to the loss function to optimize.}
\label{fig:LossTree}
\end{figure*}
\subsection{Architectural constraints}

Another approach to improving segmentation consistency with neural network predictions is through structural prior constraints \citep{trullo17,Oda2018_BESNET,Zotti2017, ELJURDI2020_BBUNET, ACNN}. For example, \cite{trullo17} introduced a collaborative architecture in order to iteratively refine the posterior probability and provide information about neighboring organs. \cite{ELJURDI2020_BBUNET} integrated location and shape prior into the learning process, by introducing bounding filters at the level of the skip-connections in a U-Net model. 
Another example is \cite{ACNN}, where the authors modified the decoder layers of a U-Net-like structure in order to incorporate prior via super resolution ground truth maps.

In many works, authors introduce both structural and loss constraints. For instance, \cite{Oda2018_BESNET} extended upon cross-entropy to introduce a boundary enhanced loss function similar to that of \cite{caliva2019_MIDL} and \cite{Arif2018}. However, instead of weighting by the errors through distance maps, \cite{Oda2018_BESNET} added an extra decoder branch to the U-Net network in order to predict hard to segment boundaries. 
In the same manner, \cite{Zotti2017} integrated the center of mass and the contour prior into their loss function, which were obtained from an encoder-decoder structure trained end-to-end, along with the segmentation network.

\section{High-level prior-based loss functions} \label{categorization}

In this section, we review the approaches that aim to integrate high-level prior for medical image segmentation at the level of the loss function, listed in \tablename~\ref{tab:lossResume}. In order to contextualize the high-level prior-based loss functions, we present a categorization of loss functions, for medical image segmentation, in Fig. \ref{fig:LossTree}: we distinguish between data-fitting losses (as described in Section \ref{lossfunctions}) that model the problem at hand and regularization losses. Prior-based losses are considered to act as regularization losses, and  can be classified according to the nature of prior that they incorporate: low-level prior (as already reviewed in Section \ref{GTtransformationlosses}) or high-level prior (the current section), which is at the heart of this review.
We have categorized the high-level priors (and subsequently the 11 reviewed papers in this section) according to the nature of the constraint: size constraint, topology, shape constraint and inter-regions constraints. We first start by the problem formulation.


\subsection{Problem formulation}\label{formulation}
In the general problem formulation of finding the segmentation network parameters by ways of optimizing a loss function, one only has the data-fitting loss, as stated in Section \ref{lossfunctions}. Training the network thus boils down to minimizing only the loss term of Eq. \ref{general}.
The integration of a prior into the loss function can be seen as a constrained optimization problem \citep{KERVADEC2019_SizeLoss, Peng2020_Discrete, Pathak2015}. Thus, in addition to the minimization of the data-fitting term $\mathcal{L}_{fit}=\sum_{p \in \Omega} L(\widehat{y}_p, y_p) $, some constraints $C_j$ to be satisfied are added. The goal is to find the network parameters $\theta$ that minimize:
\begin{equation}
     \mathcal{L} =\mathcal{L}_{Fit} \quad s.t. \quad  C_j(\widehat{y})\leq 0 \quad j : 1,...,I
   \label{general_constraint}  
\end{equation}
where $I$ is the total number of constraints in the problem. The fitting loss function $\mathcal{L}_{Fit}$ can be any of the common losses such as Dice or cross-entropy, as described in Section \ref{lossfunctions}, whereas the constraints are mathematical representations of the prior, which relate to the number of connected components, the size of the organ, the topology etc. From here on, one can distinguish between a variety of optimization and training strategies. 
 Moreover, optimization can be done either in a continuous domain where the formulated loss function is mainly derived from soft probabilities \citep{KERVADEC2019_SizeLoss, Clough2019} or a discrete domain which directly targets hard-label assignments \citep{Peng2020_Discrete}. 
One common method for solving constraint CNN training is through the method of Lagrange multipliers, also known as the penalty-based method. Such method models the constraint as a penalty term $\mathcal{L}_{penalty}$ in the loss function weighted by a parameter $\lambda$
as follows:
\begin{equation}
 \mathcal{L} = \mathcal{L}_{Fit} + \lambda \mathcal{L}_{penalty}
   \label{penalty-based-approach}  
\end{equation}
 The additional loss term must be differentiable, convex and produces a value proportional to the degree of constraint violation. The weighting factor $ \lambda $ can be either predefined throughout training (static training) or fine-tuned along training (dynamic training). We now move on to our categorization by prior nature.




\subsection{Size constraint}

The size of an organ is a feature that has a known range of variability. In \citep{KERVADEC2019_SizeLoss}, the idea is to integrate this information of the organ size information into the segmentation process, and to constraint the predicted organ area to be in this known range.
The problem is to estimate the organ size from a soft probability map. \cite{KERVADEC2019_SizeLoss} do not threshold the predicted label map, but rather estimate the area with the summation of the probabilities over the whole image domain:
\begin{equation}
   A(\widehat{\mathbf{y}}) = \sum\limits_{p \in \Omega}\widehat{y}_p
\end{equation}\label{area}
Then prior knowledge is used to impose a lower bound $a$ and a higher bound $b$ on the organ size.
A penalty loss function that integrates these bounds is proposed as follows:
\begin{equation}
\mathcal{L}_{size} = 
\begin{cases}
\big(\sum\limits_{p \in \Omega} \widehat{y}_p - a \big)^2 & \text{if } \sum\limits_{p \in \Omega}\widehat{y}_p \leq a,\\
\big(\sum\limits_{p \in \Omega} \widehat{y}_p - b \big)^2 & \text{if } \sum\limits_{p \in \Omega} \widehat{y}_p \geq b,\\
0 & \text{otherwise}.
\end{cases}
\label{eq:size-loss}
\end{equation}
Note that the proposed loss $\mathcal{L}_{size}$ is implemented in in a weakly supervised setting for cardiac segmentation. In this case prior knowledge is used in order to overcome the problem of partial label absence. 

\cite{Pathak2015} were actually the first to propose a size constraint optimized through biconvex Lagrangian dual methods. They formulate the ground truth as a latent distribution. Then, they alternate between bringing the probability distribution to be as close as possible to the ground truth distribution, given fixed model parameters on one hand, and optimize model parameters via gradient descent given known latent distribution on the other hand. In \citep{KERVADEC2019_SizeLoss}, the authors argued that the dual optimization problem is computationally intractable when applied to neural networks. As a result, it is more convenient to integrate the size prior directly at the level of the loss function, under the form of a differentiable penalty term and optimize model parameters accordingly.

Methods of \cite{KERVADEC2019_SizeLoss} and \cite{Pathak2015} explore the optimization of neural networks under constraints, given a continuous domain. However, conducting optimization from a continuous perspective may not guarantee that the discrete constraints are satisfied. The issue of the discrete nature of anatomical constraints have led to discrete optimization of neural networks in \citep{Peng2020_Discrete}, where the authors investigated the alternating direction method of multipliers algorithm (ADMM), in order to incorporate boundary smoothness and size constraints. The ADMM method is a variant of the augmented Lagrangian scheme, which allows the decoupling of the continuous optimization of neural network parameters by gradient descent, from the discrete optimization of size constraints.



\begin{figure}[t!]
\centering
\includegraphics[width=7.5cm]{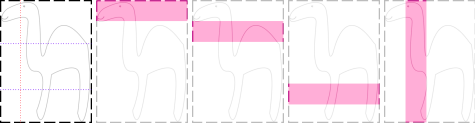} 
\caption{Tightness prior. (i)  Camel Image;  (ii), (iii), (iv) and (v) illustrate that any segment (pink stripe) of width $w$ crosses at least $w$ pixels of the camel, whether the stripe is horizontal or vertical. Figures from \citep{kervadec2019constrainedLogBarrierExtensions}. \textit{[permission pending]}}
\label{fig:camel}
\end{figure}

Still in a weakly supervised setting, \cite{kervadec2020} aim to exploit bounding box prior as means of extracting size and tightness constraints. 
With bounding box annotations, they are able to constrain the organ size, but also location, inside the bounding box \citep{kervadec2019constrainedLogBarrierExtensions}.
Finally, authors argue that the segmented region should be sufficiently close to the sides of the bounding boxes. Thus, each horizontal or vertical line that is parallel to the sides of the bounding box is to intersect the target segment at least once. As a result, the sum of pixels along the line should be greater than the sum of pixels belonging to the label (Figure \ref{fig:camel}). To integrate all three constraints, authors adopted a Lagrangian optimization method with log-barrier extensions \citep{kervadec2019constrainedLogBarrierExtensions}. The method involves introducing a standard log-barrier function \citep{boyd__2004} that evades the need for dual optimizations and their issues. Optimization under the log-barrier extensions have been introduced previously \citep{Chouzenoux2019}; however, it is still a novel research direction in medical image segmentation. 


\subsection{Topology constraints}

Topology is concerned with the properties of spatial objects by abstracting their connectivity, while ignoring their detailed form \citep{Segonne2015}.
In this section, we present work which are based on explicit topology modeling, through the use of Betti numbers, a measure of topological structures \citep{hu_NIPS2019,Clough2019}, and skeletonization \citep{clDice2020}.

Betti numbers are topological invariants. They are determined for a dimension $f$:  Betti number $\beta_f$ refers to the number of $f$-dimensional holes on a topological surface. For example, $\beta_0$ represents the number of connected components and $\beta_1$ the number of holes.
Betti numbers are discrete, obtained on thresholded binary predictions, and as such cannot be used directly for CNN training. Instead, \cite{hu_NIPS2019} and \cite{Clough2019} have exploited the notion of persistent homology. Persistent homology is a transformation that encodes the evolution of topological structures of nested spaces. In the case of prediction map, it will perform a series of thresholding on the prediction maps and summarize these information in concise format. 

\cite{hu_NIPS2019} exploit homology via persistent diagrams $(Dgm)$. Each persistent diagram contain a finite number of 2-dimensional dots ($\beta_{i,j}$) corresponding to a topological structure (or Betti number) that is born at threshold $i$ and killed at a threshold $j$. 
Then, the goal is to find the best one-to-one correspondence noted as $\gamma$, between the sets of dots corresponding to the ground truth persistent diagram ($Dgm(g)$) and to the predicted persistent diagram ($Dgm(f)$), by minimizing the squared distance between them as follows: 
\begin{equation}
\begin{aligned}
    \min_{ \gamma \in \Gamma} \sum_{\beta_{i,j} \in Dgm(f) } \Big [ birth(\beta_{i,j}) - birth(\gamma(\beta_{i,j})) \Big ]^2 \\
    \displaystyle
   +\sum_{\beta_{i,j} \in Dgm(f)} \Big [ death(\beta_{i,j}) - death(\gamma(\beta_{i,j})) \Big]^2 
   \label{Topo_PersistentHomology}
   \end{aligned}
\end{equation}
where $\Gamma$ is the set of all possible bijections from  $ Dgm(f)$ to  $Dgm(g)$.

\begin{figure}[t!]
\begin{center}
\includegraphics[width=7cm]{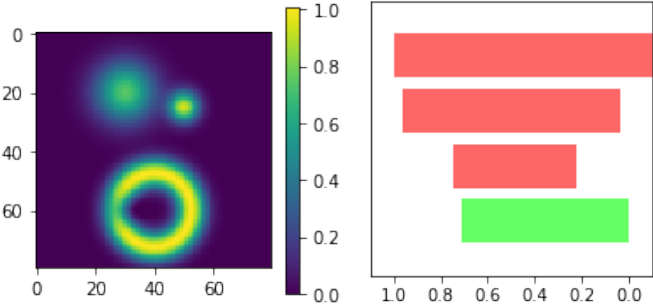} 
\caption{Persistent barcode diagram. The probability map on the left contains three visible regions of high intensity, which correspond to the three persistent 0-dimensional features shown as
red bars in the right diagram. The map also contains a loop
of high intensity, corresponding to the one persistent 1-
dimensional feature, shown here as a green bar on the
barcode diagram, with threshold values corresponding to birth and death of a topological feature on the x-axis. Figures from \citep{Clough2019}. \textit{[permission pending]}}
\label{fig:topology-barcode}  
\end{center}
\end{figure}

Instead of persistence diagrams, topological structures can be represented through persistence barcodes, as in \citep{Clough2019} (Figure \ref{fig:topology-barcode}). Longest bars are considered to have the most meaningful topological features in the data. Ideally, there is exactly one bar corresponding to each dimension $k$ of the Betti numbers, which is consistent over all thresholds.   \cite{Clough2019} consider the ground truth topological measure as a given, and compare it to the longest most persistent topological feature evident in the predicted persistence barcodes. 
Hence, their loss aims to get rid of transient topological components (shortest bars), while coaxing the network to produce probabilities that would maximize the longest bars, and is expressed as follows: 
\begin{equation}
\smallskip
\begin{aligned}
   \mathcal{L}_{Topo} = \sum_{f} \Big(  \sum_{l = 1 }^{\beta_f} \left (1- \big | birth{ \widehat{(\beta_{k,l}})}  -death({ \widehat{\beta_{k,l}}}) \big |^2 \right )\\
   \displaystyle
      +\sum_{l = \beta_f +1}^{\infty} \big | birth{ \widehat{(\beta_{k,l}})}  -death({ \widehat{\beta_{k,l}}}) \big |^2 \Big)
\label{topo_loss}
\end{aligned}
\end{equation}

Another important concept that reveals topological properties of objects is the skeleton. Skeletonization is the process of obtaining compact representations of images and objects while still preserving topological properties. The idea of \citep{clDice2020} is to constrain the skeleton of the predicted map to match the skeleton of the ground truth map. This prior is used in the segmentation of vessels and neurons in both 2D and 3D. While the skeleton of a shape can be obtained with various approaches, the authors underline that using a discrete operation such as the Euclidean distance transform does not allow to obtain a differentiable approximation. Thus, they propose to use morphological thinning, a sequence of dilations and erosions. To handle the soft probability map values, erosions and dilations are replaced by their grey-scale equivalent (min and max filters), giving rise to ‘soft-skeletonization’. In the CNN, iterative min- and max-pooling is applied as a proxy for morphological erosion and dilation. Once the skeleton is computed, an appropriate prior loss term can be designed.

Let $\mathbf{s}$ and $\mathbf{\widehat{s}}$ be the ground truth and the predicted skeleton respectively, of size $|\Omega|$. The sensitivity (or recall) between the predicted segmentation and ground truth  skeleton is introduced as $T_{sens}(s, \widehat{y}) =  {|s \cap \widehat{y}|}/{|s|}$.
 Likewise, the precision between the ground truth mask $y$ and the predicted skeleton $\widehat{s}$ is defined as: $T_{prec}(\widehat{s}, y) = {|\widehat{s} \cap y|}/{|\widehat{s}|}$
The clDice is defined as the F1-score between precision $T_{prec}$ and sensitivity $T_{sens}$ as follows:  
\begin{equation}
  \mathcal{L}_{clDice} = 2 \frac{T_{prec}(\widehat{s}, y) \, T_{sens}(s, \widehat{y})}{T_{prec}(\widehat{s}, y) + T_{sens}(s, \widehat{y}) },
  \label{skeleton-loss}
\end{equation}

Interestingly, for all three approaches \citep{hu_NIPS2019,Clough2019,clDice2020}
results show the topological prior do not fully outperform the no-prior approaches, as measured with regional metrics such as Dice, but they increase specific topological metrics, such as the clDice accuracy in \cite{clDice2020}.

%
%



\subsection{Shape constraint}
\begin{figure}[t!]
\begin{center}
\includegraphics[width=7.5cm]{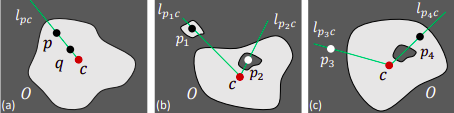}
\caption{Star shape prior. (a) Star shape object $O$ w.r.t. the supplied object center $c$ (red dot). (b) Example of star shape constraint violation. (c) Example when the star-shape prior loss is required. Figures from \citep{Mirikharaji2018} \textit{[permission pending]}.}
\label{fig:star-shape} 
\end{center}
\end{figure}

There are numerous shape descriptors, such as geometric features, moments, shape transforms, scale-space, polygonal approximation. The difficulty stems from the fact that they must be computed on real-valued maps and given their discrete nature.
Inspired by \citep{olga}, \cite{Mirikharaji2018} propose a loss that preserves the segmented region to have a shape star, for the task of segmenting dermoscopic skin lesion. An organ is said to have a star shape if, for any point $p$ inside the object, all the pixels $q$ lying on the straight-line segment connecting $p$ to the object center $c$, are inside the object (Figure \ref{fig:star-shape}).  $l_{pc}$ be the line segment connecting pixel $p$ to the object center $c$, and $q$ be any pixel incident on line $l_{pc}$. The proposed loss is expressed as:
\begin{equation}
    \mathcal{L}_{star} = \sum_{p \in \Omega} \sum_{q \in l_{pc}} B_{p,q}.
    \big|y_{p} - \widehat{y_{p}}\big|.
    \,
    \big|\widehat{y_{p}} - \widehat{y_{q}}\big| 
    \label{Star_shape_prior}
\end{equation}
where 
$$
    B_{p,q} = 
\begin{cases}
1 & \text{if }  {y_{p}} = {y_{q}}; \\
0 & \text{otherwise}. 
\end{cases}
$$
Star shape prior is a way to promote convexity for organ shapes. The star-shape loss registers significance improvement on segmentation performance, given a variety of networks such as U-Net and ResNet-DUC \citep{ResNet-DUC-HDC}.

In \citep{Dolz2017}, the idea is to enforce compactness. This property is represented by the ratio of boundary length-squared to area, that is required to be as small as possible. Here the boundary length requires to estimate a discrete form of this ratio functional, not with the usual continuous variable $\mathbf{y}$,
but with a discrete binary vector
$\mathbf{\widehat{z}} \in \{0,1\}^{|\Omega|}$:
\begin{equation}
    \mathcal{L}_{compact} = \frac{P(\mathbf{\widehat{z}})^2}{A(\mathbf{\widehat{z}})}
    \label{compact}
\end{equation}

where $A(\mathbf{\widehat{y}})$ and $P(\mathbf{\widehat{z}})$  represent the predicted organ area and perimeter respectively. Area is computed as $\sum_{p \in \Omega}\widehat{z}_p$ and perimeter is proportional to the
number of neighboring pixels with different labels, and thus computed as: $P(\mathbf{\widehat{z}}) \propto \sum_{p \in \Omega}\sum_{q \in \mathcal{N}_p}\big|\widehat{z}_{p} - \widehat{z_{q}}\big|$.

The proposed loss is dimensionless, unbiased and position independent. However, due to the discrete nature of the prior involved, optimization of this compactness-based loss comes with great challenges. For this reason, \cite{Dolz2017} alternate between optimizing the network parameters with SGD and
optimizing the discretely-constrained segmentation labels,
via ADMM. 


\subsection{Inter-regions constraint losses}

In the case of multi-label segmentation, specific interactions between regions, known a priori, can be authorized or forbidden: adjacency relations between organs are handled in \citep{Ganaye2019}, while \cite{BenTaieb2016} propose solutions to enforce regions exclusion and inclusion.

\begin{figure}[t!]
\centering
\includegraphics[width=7cm]{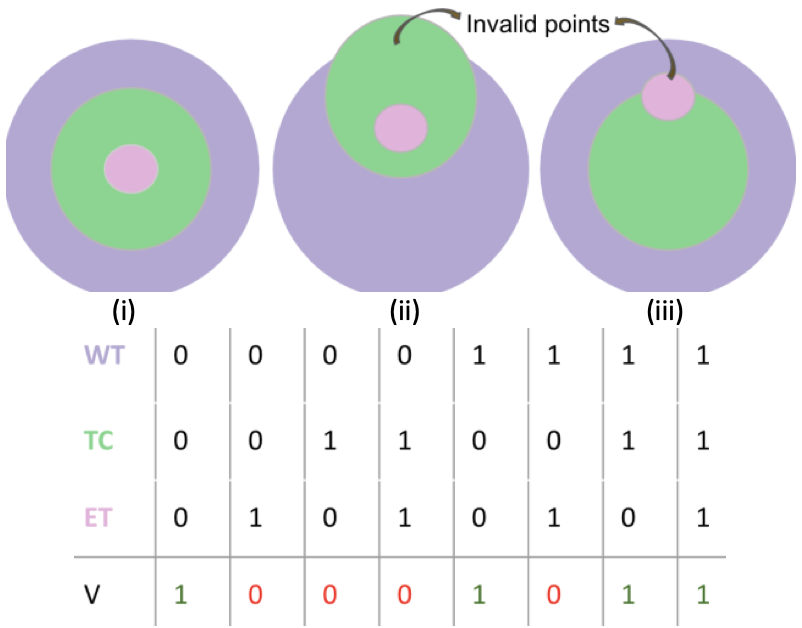} 
\caption{Inter region constraint prior. Three anatomical objects, (WT, violet), (TC, green) and (ET, pink), have $2^3=8$ possible combinations of existence. Given the correct anatomical topology specified in (i), the validity map V can then be derived for each of the 8 cases as shown in the last row of the table. Figure from \citep{Reddy2019} \textit{[permission pending]}}
\label{fig:interregion}
\end{figure}

Focusing on gland histology images, \cite{BenTaieb2016}  have identified that the cell and the object to be segmented, is made of two nested regions: one region (the lumen) is contained in another one (the epithelium). \cite{BenTaieb2016} integrated this spatial property by introducing a binary validity indicator map. A  validity indicator map $V$ returns 1 if a given label $y_p$ corresponds to a topologically-valid assignment, and zero otherwise. In this way, the network not only penalizes incorrect label assignment per pixel, but also penalize incorrect label hierarchy. Their topological loss term is based on a $V$-weighted cross-entropy, defined as follows: 

\begin{equation}
    \mathcal{L}_{Topo } = \sum_{p \in \Omega} \sum_{r = 1}^{K} -y_p^r \, \log(\widehat{y}_p^r). V
    \label{inter-1}
\end{equation}
where 
$$
V=\begin{cases}
1 & \text{if topology is in accordance;}\\ 
0 & \text{otherwise}.  \\
\end{cases}
$$

\cite{BenTaieb2016} add a boundary smoothness term that takes into consideration the difference between probabilities of pixels corresponding to same labels, to the proposed topological loss. The two constraints are combined via penalty-based optimization. The work of \cite{Reddy2019} has shown the relevance of \cite{BenTaieb2016}'s proposed topological loss for brain tumor segmentation in 3D MRI images \figureautorefname~\ref{fig:interregion}.
Note that the inclusion-exclusion loss (and the boundary smoothness loss) are not optimized alone, but are added to a fitting loss, which may be cross-entropy \citep{BenTaieb2016} or Dice \citep{Reddy2019} and the total loss function, which is the sum of all three losses, is optimized through regular stochastic gradient descent.

Both publications \cite{BenTaieb2016} and \cite{Reddy2019} validate the importance of the proposed prior loss in different tasks and modalities. However, the use of the prior loss does not compensate for the need of fitting losses such as Dice or cross-entropy. The presence of such losses is generally necessary for the convergence of the segmentation framework. Moreover, the method still depends on penalty-based optimization to balance out the two constraints, which does not accommodate the interplay and relations between the different constraints \citep{kervadec2020}.

In \cite{Ganaye2019}, the authors propose a loss that takes into consideration the relationships between neighboring anatomical objects. From the ground truth label maps, one can define an adjacency matrix $\mathbf{A}$ of general binary term $A_{ij}$ between regions, that represents whether two regions can be adjacent or not. Then the set of forbidden label connections can be defined as follows: $F=\{(i,j) | A_{ij}=0\}$.

However, an adjacency measure computed from the outputs of the CNN, which are probability maps and not label maps, is more difficult to define. When two regions $i$ and $j$ should not be connected, i.e. $(i,j) \in F$, then the probability for a pixel and its neighbors to belong to $i$ and $j$ must be close to zero. Let $\widehat{y}_p(i)$ (resp. $\widehat{y}_q(j)$) be the probability of pixel $p$ (resp. $q$) to belong to class $i$ (resp. $j$), the authors propose to model the constraint by the product $\widehat{y}_p(i)\widehat{y_q}(j)$. The adjacency measure is then:
\begin{equation}
    a_{ij} = \sum_{p \in \Omega}\sum_{q \in \mathcal{N}_p} \widehat{y}_p(i)  \widehat{y}_q(j)
    \label{ganaye0}
\end{equation}
Following this, the proposed loss consists of forcing all the forbidden adjacency relationships, with respect to the relations defined in the set of impossible transitions, $F$:
\begin{equation}
    \mathcal{L}_{adj} = \sum_{(i,j) \in F} a_{ij}
    \label{ganaye}
\end{equation}
The authors then solve the constraint optimization problem via the penalty-based method mentioned in Section \ref{formulation}. In the experiments, datasets with number of regions ranging from 20 to 135 are investigated. Interestingly, as the shape and size constraints, the proposed adjacency loss does not require the ground truth segmentation -- it just requires the definition $F$, thus the segmentation method can be evaluated in a semi-supervised framework. 

Model performance, tested using Dice, Hausdorff and Mean Surface Distance, shows no significant improvement with respect to Dice and surface metrics with respect to the established baseline. However, incorporating adjacency constraints registers considerable improvement with regards to the distance metric. These results are consistent over all datasets (in 2D and 3D) and settings, i.e. with full or semi supervision. Depending on the applications, one limitation of the approach may be the assumption that all patients have the same inter-organ connectivity.

\begin{table*}[ht!]

\begin{center}
 \caption{List of reviewed papers with respect to the category of the prior types: topology, size, shape, inter-regions priors. Evaluation metrics:  std refers to the standard evaluation metrics such as pixel-wise accuracy (pA), dice similarity coefficients (DSC), Hausdorff distance (HD); specif. means that the paper uses prior-specific metrics. SGD: stochastic gradient descent. ADMM: Alternating direction method of multipliers. }

\resizebox{0.65\paperwidth}{!}{
 \Rotatebox{90}{%
\begin{tabular}{|l|>{\centering\arraybackslash}p{2.5cm}|p{4.7cm}|>{\centering\arraybackslash}p{1cm}|c|>{\centering\arraybackslash}p{2cm}|p{2cm}|c|c|c|c|p{3.15cm}|}
\hline

Paper &Architecture  & Optimized features  & Dim. & Class & Training & \multicolumn{2}{c|}{Evaluation metrics}   & Supervision & Constraint  & Optimization  &~~Optimization \\ 
   &  &    &  & nb & strategy & ~~~~~std  &  specif &  &  nb & domain & ~~~~~~method  \\ \hline
 
\multicolumn{12}{l}{\textbf{Topology}} \\ \hline

\citep{hu_NIPS2019} 
 & 
U-Net, DIVE, Mosin 
  & Birth and death  of topological components (persistent  homology) &  
      2D  
      3D & 1& 
       static  
      with 
      refinement 
      & 
      pA &
     yes 
& full & 1 & Continuous & 
      Penalty based  
      approach  
      via SGD  \\\hline 
 \citep{Clough2019} & U-Net  & Betti number
      & 
      2D 
       3D & 1 & static & DSC & yes & semi & 1 & Continuous& 
      Penalty based 
      approach 
      via SGD 
\\ \hline 
 \citep{clDice2020} & 
      U-Net   
      FCN 
 &  Skeletons   & 
     2D 
     3D 
& 1 & static & 

     DSC, pA &
yes
& full & 1 & Continuous &  
      Penalty based  
      approach 
      via SGD \\
\hline 

\multicolumn{12}{l}{\textbf{Size }} \\ \hline 
\citep{KERVADEC2019_SizeLoss} & 
     U-net  
     E-Net  
& 
    Lower and upper bounds on organ size
  &  
     2D  
     3D 
& 1 & static& DSC & &weak  & 1-2 & Continuous &  
      Penalty based  
      approach 
      via SGD 
 \\ \hline
    \citep{kervadec2019constrainedLogBarrierExtensions} 
 & 
     U-net 
     E-Net  
& 
Combined tightness and size prior constraints
 & 2D & 1& static & DSC & - & weak & 3& Continuous & 
      Lagrangian  
       multiplier 
      with Log-barrier 
      extensions 
 \\ \hline
   \citep{Peng2020_Discrete}  
& 
   U-Net, E-Net  
      &  
     Binary  
     segmentation 
     proposals 
   and Lagrangian 
     multipliers &  
     2D 
     3D 
& 1 &N/A & DSC & - & weak &2 &Discrete  & ADMM\\ \hline

\multicolumn{12}{l}{\textbf{Shape}}  \\ \hline 
\citep{Mirikharaji2018} & 
     U-Net  
     ResNet-DUC 
&  
      Star-shapeness
     & 2D & 1 & static &
    DSC, pA, Jaccard,   Spec,
    Sens & -
& full & 1  & Continuous &  
      Penalty based  
      approach 
      via SGD 
\\ \hline
\citep{Dolz2017} & 
     FCNN 
&  
      Ratio of boundary length to area
     & 3D & 1 & N/A &
    DSC
     & -
& full & 1  & Discrete &  ADMM\\ \hline
\multicolumn{12}{l}{\textbf{Inter-regions}}  \\ \hline 
\citep{Ganaye2019} & SD-Net&
     Prohibited and permissible connectivity between organs   
&
   2D  
   2.5D 
   3D 
& 20-135 & dynamic& DSC, HD & yes & full, semi & 1 & Continuous &  
      Penalty based  
      approach 
      via SGD   \\
\hline

\citep{BenTaieb2016}  & 
     Alexnet-FCN 
     FCN-8 
     U-Net   &

     Boundary 
     smoothness and
     containment/exclusion 
     properties 
  & 
2D &  2 & static& 
     DSC, pA 
      & -& full & 2 & Continuous & 
      Penalty based  
      approach
      via SGD 
 \\ \hline
\citep{Reddy2019}  &  FCN 

&      Boundary 
     smoothness and
     containment/exclusion 
     properties  & 

2D & 3& static & 

     DSC,   
     HD 
& -
& full & 2& Continuous & 
      Penalty based 
      approach 
      via SGD 
 \\ \hline

\end{tabular}}
}
\label{tab:lossResume}    
\end{center}

\end{table*}

\section{Discussion} \label{challenges}



In addition to the common challenges in deep networks training, such as overfitting, scarcity of annotated data, class imbalance, and gradient divergence which are extensively discussed in  \citep{Hesamian2019,litjens_survey_2017,Havaei2016} for example, there are particular challenges when dealing with a prior-based term and its incorporation into the loss function. In this section, we summarize and highlight a number of key aspects of embedding a prior based loss function into a segmentation network. 

\subsection{The nature of the prior }
The high-level prior, as we defined it, expresses high-level features regarding the object of interest, with interpretable insight with respect to the organ geometry or anatomy. This prior can stem from medical knowledge (e.g. organ size range, organ connectivity) and as such, can be used in a weakly and semi-supervised learning context to improve performance. Sometimes, the prior has to rely on features extracted from the ground truth label maps, see for example the Betti numbers or the skeletonization process. In this case, its usage is restricted to full supervision.

\subsection{The challenge of soft probability maps}
One major challenge is to compute features from soft probability maps. A binary map expressing the object shape is much easier to characterize with usual shape features (e.g. circularity, compactness, isoperimetric ratio, skeleton). However, thresholding the probability map to make it binary can render the loss function non-differentiable. Some parameters can be estimated from probability maps, e.g. the predicted organ size in \citep{KERVADEC2019_SizeLoss}. Other features require to resort to a discrete optimization scheme, such as the predicted organ boundary length \citep{Dolz2017} (see Section \ref{opt-strat}).
Another way of dealing with the soft probability maps is to  impose a series of thresholds, to monitor topological changes \citep{Clough2019,Hu2019}.
However, their method is not generic and cannot be applied to all prior properties. These issues become more complicated as the type of prior handled becomes more complex, and loss functions often end up being non-convex or hard to optimize.

\subsection{Continuous vs discrete optimization strategies} \label{opt-strat}

A common and simple way to integrate constraints within a continuous domain is through the penalty based method. Penalty based method involves formalizing the constraint as an addition penalty loss term in the main loss function weighted by a parameter $ \lambda $ which may be statically or dynamically defined through training. The novel loss term which includes the main per-pixel fitting loss and the novel penalty loss are then optimized using regular stochastic gradient descent. Many works adopt the penalty-based method while dealing with anatomical prior for its ease of formulation and use \citep{clDice2020,KERVADEC2019_SizeLoss,Hu2019,Clough2019,BenTaieb2016, Mirikharaji2018}. Despite this simplicity, penalty based approaches may not guarantee constraint satisfaction. Moreover, they require careful fine-tuning of their weighting terms, which may not be convenient in the case where multiple constraints occur and where one constraint may over-shadow the others.

One way to deal with multiple constraint optimization, demonstrated in \citep{kervadec2020,kervadec2019constrainedLogBarrierExtensions}, is through Lagrangian optimization with log-barrier extensions. The method involves introducing a standard log-barrier function that avoids the need for dual optimizations and their issues \citep{boyd__2004}. The method then integrates these constraints into the log-barrier function and solves the optimization process in an unconstrained manner via stochastic gradient descent. Unlike the penalty based approach, the log-barrier approach does not yield null gradients or cause oscillations between competing constraints. It is rather characterized by stable gradients that insures training stability. 
\begin{figure*}[t!]
  \subfloat[][Topological priors in retinal vessels (left), neuronal membrane (middle), myocardium of the left ventricle (right). Top: images, bottom: ground truth.]{
	\begin{minipage}[c]{0.47\textwidth}
	   \centering
	   \includegraphics[width=7.8cm]{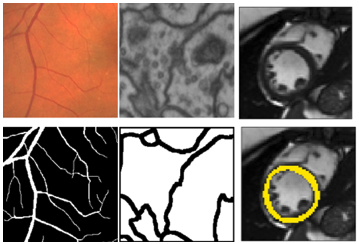} \label{fig:topo-examples} \\
	\end{minipage}
	}
 \hfill 	
  \subfloat[][Inter-region prior: adjacency constraint in brain and full body regions (top), inclusion-exclusion relationships (bottom)]{
	\begin{minipage}[c]{0.47\textwidth}
	   \centering
	   \includegraphics[width=8.5cm]{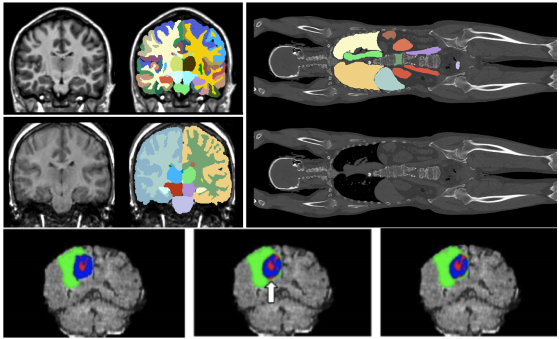} \label{fig:interregions-examples}  
	\end{minipage}}
\caption{Example of targeted segmentation objects, that can benefit from topological priors (a), inter-region priors (b). Figures from \citep{Hu2019,Clough2019,Reddy2019,ganaye2019priori}. \textit{[permission pending]}  }
\label{targetobject}

\end{figure*}
Optimization in a discrete domain can also be insightful, given the discrete nature of anatomical properties of organs. Where continuous optimization makes assumptions on the soft probabilities in order to estimate differentiable functions, discrete optimization involves extracting the features in their discrete form from model predictions, and optimizing them relative to the ground truth. One way to perform discrete optimization is through the ADMM method \citep{Peng2020_Discrete}. The ADMM algorithm generally aims at separating the optimization of the network parameters under SGD from the optimization of discrete constraint segmentation labels. Discretely optimizing networks generally benefits from the ability to solve sub-problems, either continuous or discrete, separately, and insures global optimum, which can improve solutions within a single gradient step and at higher convergence speed.

\begin{table*}[t]
    \centering
    \renewcommand{\arraystretch}{1.3}.
        \caption{Targeted segmentation objects and datasets used in the presented papers. UKb: UK Biobank. MIC12: MICCAI 2012 workshop on multi-atlas labeling. Anat3: Anatomy3. p: proprietary dataset.}

    \scriptsize
    \begin{tabular}{|lc| ccc|ccc|ccc|cc|}
    \cline{3-13}
     \multicolumn{2}{c|}{} & \multicolumn{3}{c|}{Inter-regions}& \multicolumn{3}{c|}{Topology} & \multicolumn{3}{c|}{Size} & \multicolumn{2}{c|}{Shape} \\
\cline{3-13}
         \multicolumn{2}{c|}{} & \rotatebox{90}{\citep{BenTaieb2016}}  & \rotatebox{90}{\citep{Ganaye2019}}	 & \rotatebox{90}{\citep{Reddy2019}} & \rotatebox{90}{\citep{Clough2019}} &\rotatebox{90}{\citep{hu_NIPS2019}}   & \rotatebox{90}{\citep{clDice2020}}  
         & \rotatebox{90}{\citep{Peng2020_Discrete}}
         & \rotatebox{90}{\citep{kervadec2019constrainedLogBarrierExtensions}}
         & \rotatebox{90}{\citep{KERVADEC2019_SizeLoss}}
         & \rotatebox{90}{\citep{Mirikharaji2018}}  & \rotatebox{90}{\citep{Dolz2017}}\\
         \hline
\multirow{2}{*}{CT}  & Full body & &Anat3&&&&&&&& & \\
 & Aorta/eso & &&&&&&&&& & p\\
\hline
\multirow{6}{*}{MRI} & \multirow{2}{*}{Brain} & &MIC12&BraTs&&&&&&& &\\
&  & &IBSRv2&&&&&&&& &\\
& Spine & &&&&&&&&&& \\
& Prostate & &&&&&&Promise&Promise&Promise& &\\
& Cardiac & &&&UKb &&&ACDC&&ACDC& & p\\
\hline
 \multirow{2}{*}{Photo} & Skin  & &&&&&&&&& \multirow{2}{*}{ISIC}& \\
 & lesion  & &&&&&&&&& & \\

\hline
 \multirow{2}{*}{Fundus}  & Retinal  & &&&&\multirow{2}{*}{DRIVE}&\multirow{2}{*}{DRIVE}&&&& &\\
  & vessels  & &&&&&&&&& &\\
\hline
 \multicolumn{2}{|c|}{Microscopy, }&\multirow{2}{*}{GlaS} &&&&ISBI12&&&&& &\\
  \multicolumn{2}{|c|}{histology} & & &&&ISBI13&&&&&&\\
\hline
    \end{tabular}
    \label{tab:data}
\end{table*}


\subsection{Relationship between organs and loss design}
The studied papers address various segmentation problems, as listed in Table \ref{tab:data}, that shows the targeted organs or objects to be segmented, and the datasets used in each paper. We can observe that imaging modalities are directly related with the object of interest and are mainly anatomical or tomography modalities, that exhibit a certain resolution and quality. Size and shape constraint mostly concern single instance organs, that have convex or a blob shape, such as the prostate, the cardiac ventricles, the aorta, the esophagus, or skin lesions (see for example the Promise, ACDC and ISIC datasets). Topology priors are mostly used for thin, curvilinear objects such as neuron membranes, vessels or, at a higher scale, the myocardium of the left ventricle which has a ring shape (Fig. \ref{fig:topo-examples}). Inter-regions priors can help in problems of multiclass segmentation. The adjacency constraints have shown efficiency in full body segmentation (120 regions in Anatomy3 dataset), and for multiple regions segmentation in brain segmentation (20 and 35 regions in IBSRv2 and MICCAI12 dataset). Exclusion and inclusion priors are helpful whenever there is a hierarchy in region membership. Their use is illustrated in applications, both at the microscopic and macroscopic levels. In microscopy images, cells can be composed of several layers: for example, gland cells (found in the GlaS dataset) are made of the inside region, called the glandular lumen, and the outer region, identified as the epithelial boundary. In the BraTs dataset, brain tumors in MRI are made of enhancing tumor (the deepest level), that is surrounded by the tumor core, itself surrounded by a region identified as whole tumor  (Fig. \ref{fig:interregions-examples}).

\subsection{Future trends}


The design of the prior loss is facing requirements, concerning the differentiability of the loss terms, or at least the computational complexity that must remain reasonable.
Lately, as described in this survey, some advances have been made, that explored optimization techniques such as ADMM or the barrier functions (see column 'Opt method' in Table \ref{tab:data}) and that have allowed to incorporate loss terms which are not directly optimizable by SGD. We believe future progress will originate from using advanced constraint optimization techniques, stemming from both equality or inequality constraints optimization.

Variational and optimization based approaches for image segmentation from the pre-deep learning era can also provide key pointers on how to model prior information regarding object shape and appearance. Researchers can rely on decades of works on the topic to find inspiration to design losses for their segmentation problem. In this regard, the majority of the priors are only presented in the case of binary segmentation, except for the inter-regions priors obviously. There is still work to be done to address multiple organ segmentation, with priors that are dedicated to organ shape, size or topology.

Prior based losses present promising behavior with regards to their ability to compensate for the need for full annotations, and are thus very useful in weakly and semi-supervised segmentation frameworks. This is already the case with the papers reviewed in this survey, see Table \ref{tab:data}. More generally, embedding prior information plays a role in other applications in medical imaging. For example, when using data augmentation or self-supervised approaches, we (explicitly or implicitly) make assumptions about what samples in our training set should look like, what structure they should have, and so forth. To truly make progress, we believe it is important that such assumptions are made explicit, and to discuss these methods from the perspective of more traditional methods which more heavily relied on such priors.

At last, we have noticed that often, instead of having direct effect on overall performance as measured by generic metrics such as Dice or HD, prior losses generate more plausible regions, that are hardly measurable by these generic metrics (column 'Evaluation metric' in Table \ref{tab:data}). Future trends should involve finding metrics that reflect anatomical plausibility, to show the true significance of these prior based losses.

\section{Conclusions} 
\label{conclusion}

In this paper, we presented a survey of the current state-of-the-art methods regarding high level prior based losses, for medical image segmentation. We have proposed a categorization where we grouped these methods according to the type of prior that they incorporate: size, topology, shape, and inter-region relations. We have further characterized these methods based on the type of features they optimized, the architecture they use, optimization strategy and the anatomical object that they target. We have discussed the challenges involved with the design: the fact that the prior has to be extracted from soft probability map, the optimization constraints, and the design of the loss with respect to the object of interest. Finally, we have presented some future trends that could overcome the limitations of current research works and hope they can be useful to foster research in this promising field. 

\section*{Acknowledgments}
The authors would like to acknowledge the National Council for Scientific Research of Lebanon (CNRS-L)  and the {\it Agence Fran\c caise de la Francophonie} (AUF) for granting a doctoral fellowship to Rosana El Jurdi, as well as the ANR (project APi, grant ANR-18-CE23-0014).  This work is part of the DAISI project, co-financed by the European Union with the European Regional Development Fund (ERDF) and by the Normandy Region. This research was conducted as part of a collaboration with the Eindhoven university of Technology under the support of the PHC Van Gogh project WeSmile.

\bibliographystyle{model2-names}
\bibliography{references, ref-caro}



\end{document}